\documentclass[pdflatex,sn-mathphys-num]{sn-jnl}

\usepackage{tikz}
\usetikzlibrary{arrows.meta, positioning, calc}
\usepackage{graphicx}
\graphicspath{{figs/}}
\usepackage{caption}
\usepackage{subcaption}
\usepackage{multirow}%
\usepackage{amsmath,amssymb,amsfonts}%
\usepackage{amsthm}%
\usepackage{mathrsfs}%
\usepackage[title]{appendix}%
\usepackage{xcolor}%
\usepackage{textcomp}%
\usepackage{manyfoot}%
\usepackage{booktabs}%
\usepackage{algorithm}%
\usepackage{algorithmicx}%
\usepackage{algpseudocode}%
\usepackage{listings}%
\definecolor{bfigInk}{RGB}{45,45,45}
\definecolor{bfigGen}{RGB}{33,116,79}
\definecolor{bfigGenbg}{RGB}{230,243,236}
\definecolor{bfigDisc}{RGB}{104,72,150}
\definecolor{bfigDiscbg}{RGB}{238,232,247}
\definecolor{bfigCyc}{RGB}{173,114,9}

\tikzset{
  bfigPlate/.style={draw=bfigInk!45,fill=bfigInk!4,rounded corners=2pt,
                    minimum width=2.15cm,minimum height=1.35cm,align=center,inner sep=2pt},
  bfigGen/.style={draw=bfigGen,fill=bfigGenbg,rounded corners=3pt,line width=1pt,
                  minimum width=2.7cm,minimum height=1.15cm,align=center,inner sep=3pt},
  bfigDisc/.style={draw=bfigDisc,fill=bfigDiscbg,rounded corners=3pt,line width=1pt,
                   minimum width=1.6cm,minimum height=0.9cm,align=center,inner sep=2pt},
  bfigObj/.style={draw=bfigInk!55,fill=bfigInk!4,rounded corners=3pt,line width=0.8pt,
                  align=center,inner sep=3pt},
  bfigFlow/.style={-{Latex[length=2.2mm,width=1.8mm]},draw=bfigInk,line width=1pt},
  bfigAdv/.style={-{Latex[length=1.7mm,width=1.4mm]},draw=bfigDisc,line width=0.9pt},
  bfigCyc/.style={-{Latex[length=1.9mm,width=1.6mm]},draw=bfigCyc,
                  line width=1pt,dashed,dash pattern=on 2.4pt off 1.6pt},
  bfigTag/.style={fill=white,inner sep=1pt,font=\tiny,text=bfigInk},
}

\theoremstyle{thmstyleone}%

\theoremstyle{thmstyletwo}%

\theoremstyle{thmstylethree}%

\begin{document}

\title[L0-Norm-Based CycleGAN for Brain CT-to-MR Synthesis]{L0-Norm-Based CycleGAN for Unpaired Brain CT-to-MR Synthesis}


\author[1]{\fnm{Yunxiao} \sur{Zhang}}\email{2025023307@qdu.edu.cn}

\author*[1]{\fnm{Shouqiang} \sur{Du}}\email{sqdu@qdu.edu.cn}

\author[2,3]{\fnm{Zexian} \sur{Xu}}\email{xuzexian@qdu.edu.cn}

\author[1]{\fnm{Chuanlong} \sur{Xu}}\email{2025010118@qdu.edu.cn}

\affil[1]{\orgdiv{School of Mathematics and Statistics}, \orgname{Qingdao University}, \orgaddress{\city{Qingdao}, \postcode{266071}, \country{China}}}

\affil[2]{\orgname{The Affiliated Hospital of Qingdao University}, \orgaddress{\city{Qingdao}, \postcode{266003}, \country{China}}}

\affil[3]{\orgdiv{School of Stomatology}, \orgname{Qingdao University}, \orgaddress{\city{Qingdao}, \postcode{266003}, \country{China}}}

\abstract{Computed tomography (CT) and magnetic resonance (MR) imaging provide complementary information for the brain, but acquiring both modalities is not always possible. The cycle-consistent adversarial network (CycleGAN) enables unpaired CT-to-MR synthesis, but because the standard L1-norm cycle-consistency penalty grows linearly with the residual magnitude, the network tends to focus on large residuals. This paper introduces an L0-norm-based cycle-consistency term, in which the L0-norm is approximated by a smooth and bounded function. This approximation not only increases sensitivity to small residuals but also effectively suppresses the influence of large residuals. The proposed model and an otherwise identical L1-norm baseline were trained on 102 images and evaluated on 9 test images. The proposed model achieves a lower mean squared error (0.02746 versus 0.03464, a 20.7\% reduction), a higher mean peak signal-to-noise ratio (15.80 versus 14.77 dB), and a higher mean structural similarity index (0.5689 versus 0.5407). It also has a lower maximum MSE and higher minimum PSNR and SSIM than the baseline. These results indicate that modifying the cycle-consistency penalty can improve reconstruction accuracy on this dataset.}




\maketitle

\section{Introduction}\label{sec1}

Disorders of the nervous system are among the largest contributors to ill health worldwide. In the Global Burden of Disease Study 2021, 37 neurological conditions ranked as the leading group cause of disability-adjusted life-years, affecting 3.40 billion people, or 43.1\% of the world population \cite{GBD2024}. Stroke, Alzheimer's disease, and other dementias are among the ten conditions with the highest age-standardised rates of disability-adjusted life-years \cite{GBD2024}.

Imaging is central to the diagnosis and treatment of neurological disease, and computed tomography (CT) and magnetic resonance (MR) imaging provide complementary information for the brain. MR resolves soft tissue, which is central to most assessments of the brain, whereas CT depicts bone clearly, is faster, is more widely available, and is routinely used for radiotherapy treatment planning \cite{WangT2021}. Neither modality supplies what the other does, so both are used in the diagnosis and treatment planning of diseases such as cancer \cite{WangT2021}. However, in clinical practice, imaging costs and equipment accessibility pose practical barriers to acquiring both CT and MR \cite{Wang2023}, and some patients are unable to undergo MR imaging due to metallic implants or other contraindications \cite{Shellock2004}. Consequently, many patients end up with only a brain CT and lack a brain MR. Synthesising an MR image from an existing CT image could reduce the need for an additional MR acquisition when synthetic images are acceptable for the intended task.

Supervised convolutional models are widely used in medical image analysis \cite{Litjens2017}. For CT-to-MR synthesis, representative examples include a three-dimensional fully convolutional network for CT estimation from MR \cite{Nie2016} and a SegNet-based two- and three-dimensional CNN for synthetic CT generation from T1-weighted MR \cite{Fu2019}. These supervised models require spatially aligned CT-MR image pairs from the same patient \cite{Roy2014}, which are difficult to assemble because the two examinations are acquired on different scanners, in different sessions, and with different fields of view. To overcome the bottleneck of scarce paired data, the cycle-consistent adversarial network (CycleGAN) is used to learn from unpaired datasets \cite{Zhu2017}. The network trains two generators with a cycle-consistency loss, which encourages each translation to be approximately invertible on the data support. Cycle-consistent models have since been applied to unpaired cross-modality synthesis \cite{Wang2023,Yang2020,Liu2021}. For example, Wolterink et al. \cite{Wolterink2017} applied CycleGAN to MRI and CT images to synthesize CT of the head and neck. Hiasa et al. \cite{Hiasa2018} added a gradient-consistency loss to improve boundary accuracy in pelvic MRI-to-CT synthesis.

To the best of our knowledge, current cycle-consistency losses in CycleGAN and its variants employ the L1-norm, where the penalty grows linearly with the residual magnitude. This means that the network focuses more on reducing large residuals, potentially overlooking the correction of finer details. In contrast, the L0-norm assigns equal weight to every non-zero residual. It is equally sensitive to small residuals and does not diminish its attention to them in the presence of large ones. 

Therefore this study replaces the L1-norm penalty with an L0-norm-based cycle-consistency term implemented through a smooth approximation of the L0 indicator. This forces the network to drive the error of each pixel toward zero, thereby effectively preserving reconstruction details. Moreover, the approximation makes the L0-norm differentiable and gives it a redescending influence function: the per-residual gradient is bounded and exhibits an initial rise followed by a decline as the residual increases from 0 to 1. This approximation not only increases sensitivity to small residuals but also effectively suppresses the influence of large residuals. Finally we appliy the proposed cycle-consistency term to unpaired brain CT-to-MR synthesis and evaluates it against an otherwise identical L1-norm baseline in an image-level split. The contributions are as follows.

\begin{enumerate}[1.]
  \item An L0-norm-based cycle-consistency term is introduced for unpaired brain CT-to-MR synthesis. The L0-norm is replaced by a smooth, bounded approximation that substitutes a redescending influence function for the linear weighting of the L1-norm penalty, so the loss remains sensitive to small residuals while the influence of very large residuals decays.
  \item The evaluation is performed at the image level on 111 images, with 102 images used for training and 9 images held out for testing.
  \item The comparison reports aggregate performance and the extreme-case value of each metric on the test set. The proposed model has a lower maximum MSE and higher minimum PSNR and SSIM.
\end{enumerate}

The remainder of the paper is organised as follows. Section~\ref{sec:method} sets out the problem, recalls the CycleGAN formulation, and develops the L0-norm-based cycle-consistency term together with the smooth approximation that makes it differentiable, and it then introduces the least-squares adversarial loss and states the training objective and procedure. Section~\ref{sec:exp} describes the data and split, defines the three evaluation metrics, and reports the quantitative and qualitative results, including the analysis of where in the distribution the two models differ. Section~\ref{sec:disc} explains the mechanism behind the observed pattern and discusses the interpretation of the three metrics. Section~\ref{sec:conc} concludes.

\section{Method}\label{sec:method}

\subsection{Problem setting}

Let \(X\) and \(Y\) denote the source and target domains, where
\(x \sim p_{\mathrm{data}}(X)\) denotes a brain CT image and
\(y \sim p_{\mathrm{data}}(Y)\) denotes a brain MR image. Our model aims to learn a mapping $G:X\rightarrow Y$ that produces a synthetic MR image $\hat{y}=G(x)$ from a CT image alone. Let \(I\) denote an image with \(m \times n\) pixels.

\subsection{The cycleGAN model}

The model comprises two generators and two discriminators. The generators are \(G: X \to Y\) and \(F: Y \to X\). The discriminator \(D_Y\) distinguishes between real and synthetic MR images, while \(D_X\) is trained to distinguish between real and synthetic CT images. Because \(X\) and \(Y\) are unpaired, the adversarial loss alone cannot determine which \(y\) should correspond to a given \(x\). The cycle-consistency term resolves this ambiguity by encouraging each translation to be approximately invertible, that is, \(F(G(x)) \approx x\) and \(G(F(y)) \approx y\). In the original CycleGAN, both directions are measured with the L1-norm penalty \cite{Zhu2017}:
\begin{equation}\label{eq:cyc-l1}
\begin{split}
\mathcal{L}_{\mathrm{cyc}}(G,F)
  & = \mathbb{E}_{x\sim p_{\mathrm{data}}(X)}\left[\frac{1}{mn}\lVert F(G(x))-x\rVert_1\right] \\
  & \quad + \mathbb{E}_{y\sim p_{\mathrm{data}}(Y)}\left[\frac{1}{mn}\lVert G(F(y))-y\rVert_1\right],
\end{split}
\end{equation}
And the adversarial losses are as follows \cite{Goodfellow2014}:
\begin{equation}\label{eq:gan-g}
\begin{split}
\mathcal{L}_{\mathrm{GAN}}(G,D_Y,X,Y)
  & = \mathbb{E}_{y\sim p_{\mathrm{data}}(Y)}\left[\log D_Y(y)\right] \\
  & \quad + \mathbb{E}_{x\sim p_{\mathrm{data}}(X)}\left[\log\left(1-D_Y(G(x))\right)\right],
\end{split}
\end{equation}
\begin{equation}\label{eq:gan-f}
\begin{split}
\mathcal{L}_{\mathrm{GAN}}(F,D_X,Y,X)
  & = \mathbb{E}_{x\sim p_{\mathrm{data}}(X)}\left[\log D_X(x)\right] \\
  & \quad + \mathbb{E}_{y\sim p_{\mathrm{data}}(Y)}\left[\log\left(1-D_X(F(y))\right)\right].
\end{split}
\end{equation}
The full objective is a weighted combination of \eqref{eq:cyc-l1}, \eqref{eq:gan-g}, and \eqref{eq:gan-f}. Training is performed by solving the following minimax problem:
\begin{equation*}\label{eq:minimax-log}
\arg\min_{G,F}\ \max_{D_X,D_Y}\
\Big[\mathcal{L}_{\mathrm{GAN}}(G,D_Y,X,Y)
      +\mathcal{L}_{\mathrm{GAN}}(F,D_X,Y,X)
      +\lambda\,\mathcal{L}_{\mathrm{cyc}}(G,F)\Big],
\end{equation*}
where $\lambda$ is a scalar that weights the cycle-consistency term relative to the adversarial terms. 

\subsection{L0-norm cycle-consistency penalty}\label{sec:l0}

For an image \(I \in \mathbb{R}^{m \times n}\), the L0-norm is the number of non-zero entries:
\begin{equation}\label{eq:l0}
\lVert I\rVert_0=\sum_{i=1}^{m}\sum_{j=1}^{n}\delta(I_{ij}),
\qquad
\delta(I_{ij})=
\begin{cases}
1, & I_{ij}\neq 0,\\
0, & I_{ij}=0,
\end{cases}
\end{equation}
where \(I_{ij}\) denotes the pixel value at the \((i, j)\)-th position of image \(I\).

Then, we replace the L1-norm term in \eqref{eq:cyc-l1} with the L0-norm to obtain the new cycle-consistency loss:
\begin{equation}\label{eq:cyc-l0}
\begin{split}
\mathcal{L}_{\mathrm{cyc}}(G,F)
  & = \mathbb{E}_{x\sim p_{\mathrm{data}}(X)}\left[\frac{1}{mn}\lVert F(G(x))-x\rVert_0\right] \\
  & \quad + \mathbb{E}_{y\sim p_{\mathrm{data}}(Y)}\left[\frac{1}{mn}\lVert G(F(y))-y\rVert_0\right].
\end{split}
\end{equation}
We follow the common convention of referring to this quantity as the L0-norm, although strictly speaking it is a cardinality function rather than a true norm.

The Eq. \eqref{eq:cyc-l0} differs from the L1-norm in how it weights residuals of different size. Under the L1-norm penalty, a residual of size \(r\) contributes \(|r|\) to the loss and induces a gradient of magnitude one, so the loss value grows linearly with residual magnitude. Under the L0-norm, every non-zero residual contributes one; this increases the sensitivity to small residuals and shifts the generator's objective to minimizing the number of mismatching pixels.

\subsection{Smooth approximation and gradient properties}\label{sec:approx}

Since the indicator function \(\delta(\cdot)\) is discontinuous at the origin and has zero derivative elsewhere, it provides no useful gradient information. Therefore, Eq. \eqref{eq:cyc-l0} cannot be directly optimized by gradient descent. To address this, we adopt a smooth approximation \(f_{\sigma}(\cdot)\) \cite{Zhao2014} that retains the intended L0-norm properties while remaining differentiable. This function is defined as
\[
f_{\sigma}(I_{ij}) = \tanh \left( \frac{I_{ij}^2}{2\sigma^2} \right) = \frac{e^{\frac{I_{ij}^2}{2\sigma^2}} - e^{-\frac{I_{ij}^2}{2\sigma^2}}}{e^{\frac{I_{ij}^2}{2\sigma^2}} + e^{-\frac{I_{ij}^2}{2\sigma^2}}}.
\]
where \(\sigma\) is a parameter. It satisfies \(\lim_{\sigma \to 0} f_{\sigma}(I_{ij}) = \delta(I_{ij})\).

The derivative of \(f_\sigma(I_{ij})\) is given by:

\[
f'_\sigma(I_{ij}) = \frac{I_{ij}}{\sigma^2} \operatorname{sech}^2\left(\frac{I_{ij}^2}{2\sigma^2}\right) = \frac{I_{ij}}{\sigma^2}\left(1 - f_\sigma(I_{ij})^2\right).
\]

By introducing the variable substitution \(I_{ij} = \sigma s\), we can separate the scale parameter \(\sigma\) from the inherent profile of the gradient function. The absolute value of the gradient can be expressed as \(|f'_\sigma(I_{ij})| = \sigma^{-1}g(s)\), where the shape function is defined as \(g(s) = s\operatorname{sech}^2(s^2/2)\). Its derivative is written as

\[
g'(s) = \operatorname{sech}^2(s^2/2)\left[1 - 2s^2 \tanh(s^2/2)\right].
\]

To find the maximum value of the gradient, we set \(g'(s) = 0\) and solve for the extremum point \(s^* \approx 1.02\). Substituting this result yields the upper bound of gradient magnitude:

\begin{equation}\label{eq:grad-max}
\max_{I_{ij}} |f'_\sigma(I_{ij})| = \frac{g(s^*)}{\sigma} \approx \frac{0.787}{\sigma},
\end{equation}
and this peak gradient occurs when the residual satisfies \(|I_{ij}| \approx 1.02\sigma\).

From Eq. \eqref{eq:grad-max}, the upper bound of the gradient is \(0.787/\sigma\). As \(\sigma \to 0\), this upper bound tends to infinity, the function becomes increasingly stiff, and training is highly prone to gradient explosion, making convergence difficult. Therefore, \(\sigma\) cannot be allowed to actually approach zero during training. In addition, the gradient of the approximate L0 loss first increases and then decreases, peaking at \(|I_{ij}| \approx 1.02\sigma\). This indicates that the loss is most sensitive to residuals with magnitudes close to \(\sigma\). Given that our images are normalized to the range \([-1, 1]\), \(\sigma\) confines the transition region of the loss response within a reasonable range of residuals. Therefore, an appropriate value of \(\sigma\) should be selected to match the dynamic range of image residuals.

Therefore, the actual L0-norm based cycle-consistency loss is given by:

\begin{equation}\label{eq:cyc-approx}
\begin{split}
\mathcal{L}^{(\sigma)}_{\mathrm{cyc}}(G, F) = & \mathbb{E}_{x \sim p_{\mathrm{data}}(X)} \left[ \frac{1}{mn} \sum_{i=1}^{m} \sum_{j=1}^{n} f_\sigma\left( (F(G(x)) - x)_{ij} \right) \right] \\
&+ \mathbb{E}_{y \sim p_{\mathrm{data}}(Y)} \left[ \frac{1}{mn} \sum_{i=1}^{m} \sum_{j=1}^{n} f_\sigma\left( (G(F(y)) - y)_{ij} \right) \right].
\end{split}
\end{equation}

\subsection{Least-squares adversarial loss}\label{sec:lsgan}

This paper adopts the least squares adversarial loss (LSGAN) as the adversarial constraint between the generator and the discriminator. This loss can
alleviate the vanishing gradient problem of the original logarithmic GAN
loss and provides stronger training stability \cite{Mao2017}.
\begin{equation}\label{eq:lsgan-g}
\begin{split}
\mathcal{L}_{\mathrm{LSGAN}}(G,D_Y)
  & = \frac{1}{2}\,\mathbb{E}_{y\sim p_{\mathrm{data}}(Y)}
      \left[\left(D_Y(y)-1\right)^{2}\right] \\
  & \quad + \frac{1}{2}\,\mathbb{E}_{x\sim p_{\mathrm{data}}(X)}
      \left[D_Y(G(x))^{2}\right],
\end{split}
\end{equation}
\begin{equation}\label{eq:lsgan-f}
\begin{split}
\mathcal{L}_{\mathrm{LSGAN}}(F,D_X)
  & = \frac{1}{2}\,\mathbb{E}_{x\sim p_{\mathrm{data}}(X)}
      \left[\left(D_X(x)-1\right)^{2}\right] \\
  & \quad + \frac{1}{2}\,\mathbb{E}_{y\sim p_{\mathrm{data}}(Y)}
      \left[D_X(F(y))^{2}\right].
\end{split}
\end{equation}

\subsection{Full objective and optimisation}

Combining Sections 2.1 to 2.5, our final objective function is given by:

\[
\arg\min_{G,F} \max_{D_X, D_Y} \left[ \mathcal{L}_{\text{LSGAN}}(G, D_Y) + \mathcal{L}_{\text{LSGAN}}(F, D_X) + \lambda \mathcal{L}_{\text{cyc}}^{(\sigma)}(G, F) \right].
\]

The generators follow the CycleGAN residual architecture: a \(7 \times 7\) convolution with 32 filters, two \(3 \times 3\) convolutions with stride two and 64 and 128 filters, nine residual blocks with 128 filters, two transposed convolutions with 64 and 32 filters, and a final \(7 \times 7\) convolution to three channels, followed by a tanh output and a skip connection that adds the input of the generator to its output. Instance normalisation and ReLU are used throughout the generators.

The discriminators are \(70 \times 70\) PatchGANs, that is, patch-based discriminators, each consisting of five \(4 \times 4\) convolutions with 64, 128, 256, 512 and one output channel, the first three with stride two, leaky ReLU activations of slope 0.2, and a linear output layer. The discriminator is trained to output 1 for real images and 0 for synthetic images, while the generator aims to make the discriminator output close to 1 for generated images. When updating the discriminator, samples are drawn from a pool of fifty previously generated images rather than images produced in the current iteration. All comparison experiments in this paper adopt identical adversarial loss, network architecture, training schedule and cycle weight \(\lambda\). Only the cycle-consistency penalty is modified to guarantee a fair comparison.

Both models were trained for 200 epochs with the Adam optimiser \cite{Kingma2015} at an initial learning rate of \(2 \times 10^{-4}\) with \(\beta_1 = 0.5\), a batch size of one image, and a pool of fifty previously generated images for the discriminator updates. Images were resized to \(286 \times 286\) pixels, randomly cropped to \(256 \times 256\) during training, and flipped horizontally at random. Intensities were scaled to \([-1, 1]\). The cycle weight was \(\lambda = 10\) and the approximation scale was \(\sigma = 0.1\). The baseline is the same model with an L1-norm penalty in place of the approximation, trained on the same data, architecture, adversarial loss, and schedule, so that the two models differ only in the cycle-consistency term within a common training configuration.

A schematic diagram of the final network is shown in Figure 1. The training steps are summarized in Algorithm 1.

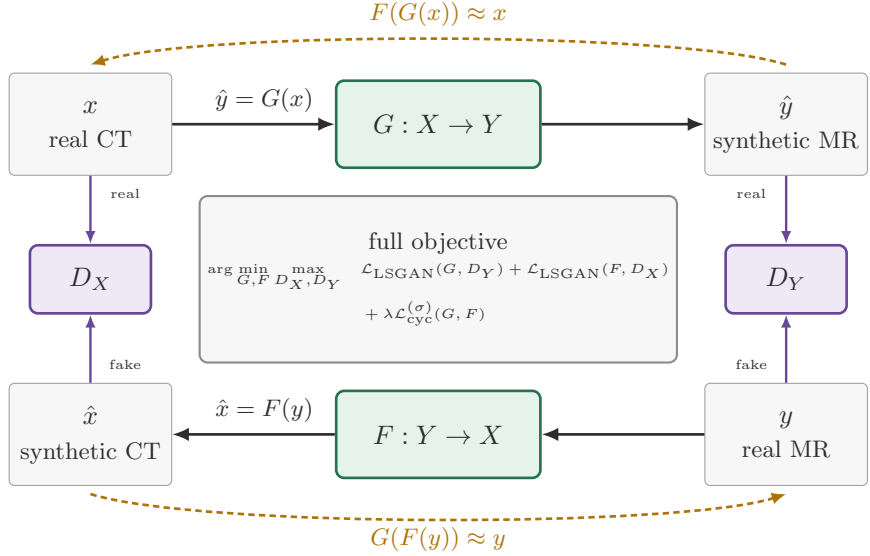
\begin{figure*}[htbp]
\centering
\begin{tikzpicture}[x=1cm,y=1cm,every node/.style={text=bfigInk}]
  \path[use as bounding box] (0.15,0.50) rectangle (11.85,7.90);

  \node[bfigPlate] (nx)  at (1.40,6.20) {$x$\\[1pt]\scriptsize real CT};
  \node[bfigPlate] (nyh) at (10.60,6.20) {$\hat{y}$\\[1pt]\scriptsize synthetic MR};
  \node[bfigPlate] (nxh) at (1.40,2.10) {$\hat{x}$\\[1pt]\scriptsize synthetic CT};
  \node[bfigPlate] (ny)  at (10.60,2.10) {$y$\\[1pt]\scriptsize real MR};

  \node[bfigGen] (nG) at (6.00,6.20) {$G:X\rightarrow Y$};
  \node[bfigGen] (nF) at (6.00,2.10) {$F:Y\rightarrow X$};

  \node[bfigDisc] (nDX) at (1.40,4.15) {$D_X$};
  \node[bfigDisc] (nDY) at (10.60,4.15) {$D_Y$};

  \draw[bfigFlow] (nx.east)  -- (nG.west);
  \draw[bfigFlow] (nG.east)  -- (nyh.west);
  \draw[bfigFlow] (ny.west)  -- (nF.east);
  \draw[bfigFlow] (nF.west)  -- (nxh.east);
  \node[font=\scriptsize,above] at (3.70,6.24) {$\hat{y}=G(x)$};
  \node[font=\scriptsize,above] at (3.70,2.14) {$\hat{x}=F(y)$};

  \draw[bfigAdv] (nx.south)  -- (nDX.north);
  \draw[bfigAdv] (nxh.north) -- (nDX.south);
  \draw[bfigAdv] (nyh.south) -- (nDY.north);
  \draw[bfigAdv] (ny.north)  -- (nDY.south);
  \node[bfigTag] at (1.86,5.28) {real};
  \node[bfigTag] at (1.86,3.02) {fake};
  \node[bfigTag] at (10.14,5.28) {real};
  \node[bfigTag] at (10.14,3.02) {fake};

  \node[bfigObj,minimum width=5.7cm,minimum height=2.2cm] (nObj) at (6.00,4.15) {%
    \scriptsize full objective\\[2pt]%
    {\tiny \(\begin{aligned}
    \arg\min_{G,F} \max_{D_X, D_Y} \quad & \mathcal{L}_{\text{LSGAN}}(G, D_Y) + \mathcal{L}_{\text{LSGAN}}(F, D_X) \\
    & + \lambda \mathcal{L}_{\text{cyc}}^{(\sigma)}(G, F)
    \end{aligned}\)}%
};

  \draw[bfigCyc] ([yshift=3pt]nyh.north)
        .. controls +(-1.4,0.45) and +(1.4,0.45) .. ([yshift=3pt]nx.north);
  \draw[bfigCyc] ([yshift=-3pt]nxh.south)
        .. controls +(1.4,-0.45) and +(-1.4,-0.45) .. ([yshift=-3pt]ny.south);
  \node[font=\scriptsize,text=bfigCyc,fill=white,inner sep=1.5pt]
       at (6.00,7.66) {$F(G(x))\approx x$};
  \node[font=\scriptsize,text=bfigCyc,fill=white,inner sep=1.5pt]
       at (6.00,0.72) {$G(F(y))\approx y$};

\end{tikzpicture}
\caption{Structure of the proposed L0-norm-based CycleGAN framework.}
\label{fig:model}
\end{figure*}

\begin{algorithm}
\caption{Training procedure of the proposed L0-norm based CycleGAN framework}
\begin{algorithmic}[1]
\Require Image pool size \(P = 50\). Generators \(G,F\); Discriminators \(D_X,D_Y\).
\State Initialize parameters of \(G,F,D_X,D_Y\); Initialize empty image pools \(\mathcal{P}_X,\mathcal{P}_Y\).
\For{each epoch}
    \For{each mini-batch with batch size = 1}
        \State Sample real images \(x \in X\), \(y \in Y\).
        \State Generate fake images: \(\hat{y} = G(x)\) and \(\hat{x} = F(y)\).
        \State \textbf{Step 1: Update Generator \(G\)}
        \State Compute loss for \(G\): Eq. \eqref{eq:lsgan-g} + \(\lambda\) Eq. \eqref{eq:cyc-approx}.
        \State Optimize parameters of \(G\) via Adam; push \(\hat{y}\) into pool \(\mathcal{P}_Y\).
        \State \textbf{Step 2: Update Discriminator \(D_Y\)}
        \State Sample a historical image \(\hat{y}_{\text{pool}}\) from \(\mathcal{P}_Y\).
        \State Compute LSGAN loss for \(D_Y\) (Eq. \eqref{eq:lsgan-g}) using real \(y\) and fake \(\hat{y}_{\text{pool}}\).
        \State Optimize parameters of \(D_Y\) via Adam (parameters of \(G\) are frozen).
        \State \textbf{Step 3: Update Generator \(F\)}
        \State Compute loss for \(F\): Eq. \eqref{eq:lsgan-f} + \(\lambda\) Eq. \eqref{eq:cyc-approx}.
        \State Optimize parameters of \(F\) via Adam; push \(\hat{x}\) into pool \(\mathcal{P}_X\).
        \State \textbf{Step 4: Update Discriminator \(D_X\)}
        \State Sample a historical image \(\hat{x}_{\text{pool}}\) from \(\mathcal{P}_X\).
        \State Compute LSGAN loss for \(D_X\) (Eq. \eqref{eq:lsgan-f}) using real \(x\) and fake \(\hat{x}_{\text{pool}}\).
        \State Optimize parameters of \(D_X\) via Adam (parameters of \(F\) are frozen).
    \EndFor
\EndFor
\State \textbf{Pool Rule:} Before pool is full, save new fakes directly. Once full, replace a stored sample with the new fake with probability 0.5.
\end{algorithmic}
\end{algorithm}

\section{Experiments}\label{sec:exp}

\subsection{Data and split}

The evaluation in this study relies on the publicly available dataset introduced by Han \cite{Han2017}, the dataset can be downloaded from \url{https://github.com/ChengBinJin/MRI-to-CT-DCNN-TensorFlow}. This dataset contains 367 CT and MR images of size 512 \(\times\) 256. We cropped each image into two 256 \(\times\) 256 images of the same modality (CT or MR). In the original CT images, white curvilinear artifacts were visible at the bottom of both the left and right sides; these artifacts were manually removed. In addition, the dataset contains CT and MRI slices from various angles and anatomical regions, which substantially increases the difficulty of training the generator and exposes it to the risk of mode collapse. To this end, we selected 111 paired images with similar structures for each modality, of which 102 were used for training and 9 for testing.

\subsection{Evaluation metrics}

Three pixel-level metrics are reported. Let $G(x)$ denote the synthetic image and $y$ the reference image, both of size $m\times n$. The mean squared error (MSE) \cite{Wang2023} is
\begin{equation*}\label{eq:mse}
\mathrm{MSE}=\frac{1}{mn}\sum_{i=1}^{m}\sum_{j=1}^{n}
\left(y_{ij}-\left(G(x)\right)_{ij}\right)^{2},
\end{equation*}
the peak signal-to-noise ratio (PSNR) \cite{Wang2023} is
\begin{equation*}\label{eq:psnr}
\mathrm{PSNR}=10\log_{10}\!\left(\frac{\mathrm{MAX}^{2}}{\mathrm{MSE}}\right)
\quad\text{(dB)},
\end{equation*}
and the structural similarity index (SSIM) \cite{Wang2004} is
\begin{equation*}\label{eq:ssim}
\mathrm{SSIM}\!\left(G(x),y\right)
=l\!\left(G(x),y\right)^{\alpha}\,
c\!\left(G(x),y\right)^{\beta}\,
s\!\left(G(x),y\right)^{\gamma},
\end{equation*}
in which $l$, $c$ and $s$ compare luminance, contrast and structure respectively. A lower MSE, a higher PSNR and an SSIM closer to one each indicate closer agreement with the reference image.

\subsection{Quantitative results}

Table~\ref{tab:metrics} reports the two models on the test set. The proposed model achieved
a lower mean MSE, a higher mean PSNR, and a higher mean SSIM than the
L1-norm baseline. The mean MSE was $20.7\%$ lower, and the mean SSIM
was higher by 0.0282, representing a $5.2\%$ relative increase.

\begin{table*}[http]
\centering
\caption{Test-set performance of the L1-norm baseline and the proposed L0-norm-based model. Values are mean $\pm$ standard deviation, with the minimum and maximum of the same statistic over the test set in parentheses. The images were scaled to $[-1,1]$, so MSE is expressed in squared normalised intensity units.}
\label{tab:metrics}
\begin{tabular}{l c c c}
\toprule
Model & MSE & PSNR (dB) & SSIM \\
\midrule
Baseline (L1-norm)      & $0.03464\pm0.00943$ & $14.77\pm1.23$ & $0.5407\pm0.0869$ \\
                   & $(0.01720,\,0.04770)$ & $(13.22,\,17.64)$ & $(0.3599,\,0.6850)$ \\
L0-norm (proposed) & $0.02746\pm0.00798$ & $15.80\pm1.33$ & $0.5689\pm0.0796$ \\
                   & $(0.01726,\,0.04048)$ & $(13.93,\,17.63)$ & $(0.3776,\,0.6419)$ \\
\bottomrule
\end{tabular}
\end{table*}

The change in PSNR is consistent with the change in MSE. The reduction in MSE from $0.03464$ to $0.02746$ corresponds to a gain of $10\log_{10}(0.03464/0.02746)=1.01$~dB, close to the reported mean gain of $1.03$~dB. The small difference reflects the fact that the mean of the per-case PSNR is not equal to the PSNR of the mean MSE.

\subsection{Extreme-case performance}\label{sec:tail}

The aggregate means in Table~\ref{tab:metrics} summarise overall performance. The extrema in the same table describe the worst and best observed values across the test cases for each metric.

On the test set, the proposed model lowered the maximum MSE from $0.04770$ to $0.04048$ ($15.1\%$), raised the minimum PSNR from $13.22$ to $13.93$~dB, and raised the minimum SSIM from $0.3599$ to $0.3776$ ($4.9\%$). The baseline retained a higher maximum SSIM ($0.6850$ versus $0.6419$). These observations indicate that the proposed model achieves more favorable extreme values for each metric.

\subsection{Qualitative comparison}

\begin{figure*}[htbp]
  \centering
  \begin{subfigure}[b]{0.235\textwidth}\centering
    \includegraphics[width=\textwidth]{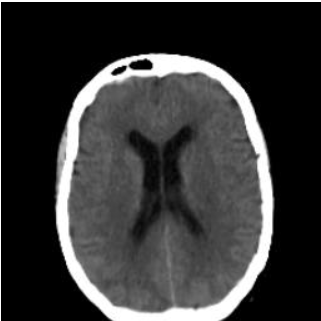}
  \end{subfigure}\hfill
  \begin{subfigure}[b]{0.235\textwidth}\centering
    \includegraphics[width=\textwidth]{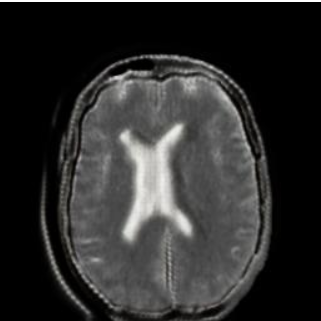}
  \end{subfigure}\hfill
  \begin{subfigure}[b]{0.235\textwidth}\centering
    \includegraphics[width=\textwidth]{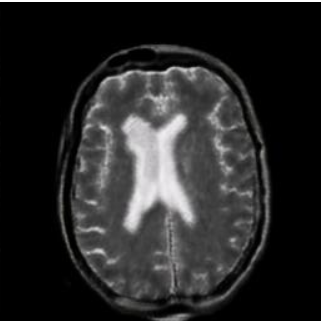}
  \end{subfigure}\hfill
  \begin{subfigure}[b]{0.235\textwidth}\centering
    \includegraphics[width=\textwidth]{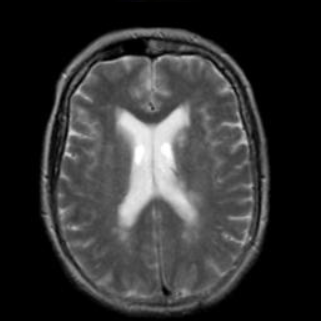}
  \end{subfigure}

  \vspace{3pt}
  \begin{subfigure}[b]{0.235\textwidth}\centering
    \includegraphics[width=\textwidth]{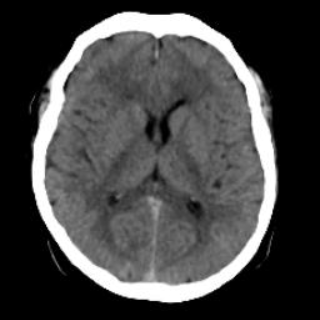}
  \end{subfigure}\hfill
  \begin{subfigure}[b]{0.235\textwidth}\centering
    \includegraphics[width=\textwidth]{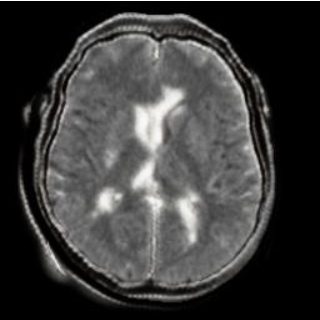}
  \end{subfigure}\hfill
  \begin{subfigure}[b]{0.235\textwidth}\centering
    \includegraphics[width=\textwidth]{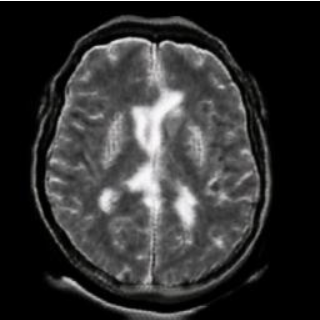}
  \end{subfigure}\hfill
  \begin{subfigure}[b]{0.235\textwidth}\centering
    \includegraphics[width=\textwidth]{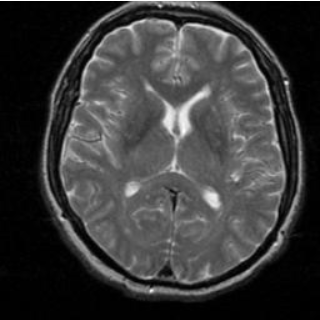}
  \end{subfigure}

  \caption{Two representative cases, one per row. Columns, from left to right: input CT, synthetic MR of the proposed model, synthetic MR of the L1-norm baseline, reference MR.}
  \label{fig:cases}
\end{figure*}

Figure~\ref{fig:cases} presents a qualitative comparison of the synthesized results. Compared with the original CycleGAN, the proposed L0-norm model is visually
closer to the reference MR image. Specifically, the proposed L0 model preserves sharper tissue boundaries while suppressing unnatural artifacts.

\section{Discussion}\label{sec:disc}

Taken together, the test results show that replacing the L1-norm cycle penalty with the approximation of the L0-norm improves mean reconstruction accuracy and the extreme-case value of each metric in this dataset. The CycleGAN architecture and the other training components are shared by the two models, so the observed differences can be examined under a common training configuration.

\subsection{Interpretation of the observed improvement}

The lower maximum MSE and higher minimum PSNR and SSIM reported in Section \ref{sec:tail} are consistent with the design of the approximation. Under the L1-norm penalty, every non-zero residual has a gradient of magnitude one. Under the approximation, the gradient is bounded, peaks at \(|I_{ij}| \approx 1.02\sigma\), and decays for larger residuals. The latter therefore reduces the relative influence of very large residuals, while retaining sensitivity near the transition band.

What's more, the scale parameter \(\sigma\) governs the location of the transition region on the gradient curve of the proposed approximation. For \(\sigma = 0.1\), the gradient reaches its maximum value of \(7.87\) at \(|I_{ij}| = 0.102\). The gradient of the approximate loss remains larger than that of the L1-norm penalty for residuals below approximately \(0.21\), and decays afterwards to about \(0.19\) at residual \(0.25\) and \(0.015\) at residual \(0.3\). The root-mean-square reconstruction error for the synthetic reference is \(0.166\), i.e., \(1.66\sigma\), which indicates that the typical scale of reconstruction errors on this dataset lies near this transition band and facilitates the desirable residual-modulating property of the loss function. It should be noted that the cycle residual is not identical to the reconstruction residual, so the above matching is only qualitative. The overall balance of the cycle-consistency constraint is jointly controlled by the hyperparameters \(\lambda\) and \(\sigma\).

\subsection{Relation to previous work and metric interpretation}

Most existing improvements to CycleGAN focus on network architecture design or introduce additional loss terms while retaining the original cycle-consistency penalty. The core contribution of this paper lies in directly modifying the cycle-consistency loss itself. Consequently, the proposed scheme can be readily transferred to other existing CycleGAN variants.

MSE and PSNR measure global error and are closely related for a fixed dynamic range, whereas SSIM compares local luminance, contrast, and structure. Agreement between the mean MSE and mean SSIM therefore supports the observed improvement.

\section{Conclusion}\label{sec:conc}

This paper introduced an L0-norm-based cycle-consistency term, implemented through a smooth, bounded approximation, and applied it to unpaired brain CT-to-MR synthesis. The approximation provides a redescending influence function: its per-residual gradient is bounded, peaks at residuals of order \(\sigma\), and decays for much larger residuals. This design offers a principled alternative to the linear L1-norm penalty, which weights the loss by residual magnitude.

After training on 102 images and holding out the remaining 9, the proposed model achieved a lower mean MSE (0.02746 versus 0.03464), a higher mean PSNR (15.80 versus 14.77 dB), and a higher mean SSIM (0.5689 versus 0.5407) on the test set. It also produced a lower maximum MSE and higher minimum PSNR and SSIM.

These results show that changing the cycle-consistency penalty can improve reconstruction accuracy in this dataset. Future work will incorporate a systematic \((\sigma, \lambda)\) sensitivity analysis, multi-centre data, and downstream task evaluation.

\section*{Data Availability Statement}

The datasets analyzed during the current study were originally introduced in the article by Han \cite{Han2017} and are available in the GitHub repository at \url{https://github.com/ChenBinJin/MRI-to-CT-DCNN-TensorFlow} \cite{Github}. This toy dataset consists of 367 paired images and is accessible via the download link provided in the repository's documentation.

\section*{Declarations}

\begin{itemize}
\item This work is supported by the National Natural Science Foundation of China (Grant No.~12371304).
\item The authors declare no competing interests.
\end{itemize}

\end{document}